\def\year{2021}\relax
\documentclass[letterpaper]{article} 
\usepackage{aaai21}  
\usepackage{times}  
\usepackage{helvet} 
\usepackage{courier}  
\usepackage[hyphens]{url}  
\usepackage{graphicx} 
\usepackage{natbib}  
\usepackage{caption} 
\usepackage{amsmath}
\usepackage{ntheorem}
\newtheorem{hyp}{Hypothesis}
\title{Explaining Reward Functions to Humans for Better Human-Robot Collaboration}
\author{
    Lindsay Sanneman and Julie A. Shah
    \\
}
\affiliations{
    CSAIL, Massachusetts Institute of Technology\\

    {lindsays, julie\_a\_shah}@csail.mit.edu

}

\begin{document}

\maketitle

\begin{abstract}
Explainable AI techniques that describe agent reward functions can enhance human-robot collaboration in a variety of settings. One context where human understanding of agent reward functions is particularly beneficial is in the value alignment setting.  In the value alignment context, an agent aims to infer a human's reward function through interaction so that it can assist the human with their tasks. If the human can understand where gaps exist in the agent’s reward understanding, they will be able to teach more efficiently and effectively, leading to quicker human-agent team performance improvements. In order to support human collaborators in the value alignment setting and similar contexts, it is first important to understand the effectiveness of different reward explanations techniques in a variety of domains. In this paper, we introduce a categorization of information modalities for reward explanation techniques, propose a suite of assessment techniques for human reward understanding, and introduce four axes of domain complexity. We then propose an experiment to study the relative efficacy of a broad set of reward explanation techniques covering multiple modalities of information in a set of domains of varying complexity.

\end{abstract}

\section{Introduction}

\begin{figure}[t]
      \centering
      \includegraphics[width=0.8\columnwidth]{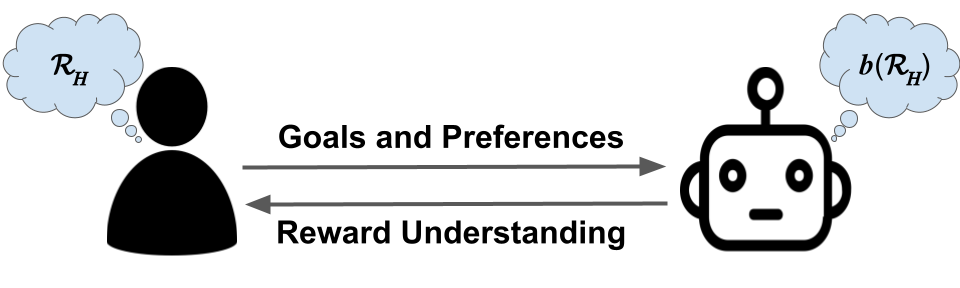}
      \caption{Bidirectional Communication for Value Alignment}
      \label{twoway}
    \end{figure}

Effective explainable AI systems are critical to constructive human-robot collaboration in a variety of settings. Humans performing separate roles in distinct contexts require different types of information about their AI teammates in order to effectively perform their tasks \cite{sanneman2020situation}. One context in which explainable AI is particularly important is the value alignment setting in which a human and an autonomous agent work together to maximize some reward, but only the human knows the true reward function \cite{fisac2020pragmatic, hadfield2016cooperative}. Through interaction, the agent infers the reward in order to become a better collaborator. It is ideal for the human to behave ``pedagogically'' in this circumstance, taking actions that best teach the agent about the true reward,  but this requires the human to track the agent's beliefs about the reward over the course of the interaction \cite{fisac2020pragmatic}. This might be difficult or intractable in some settings given human cognitive limitations. Therefore, enabling an agent to provide feedback to a human about its current understanding of the reward function, as depicted in Figure \ref{twoway}, could be of value.

In order to understand how to best explain agent reward functions in a value alignment setting, it is first important to understand which approaches for explaining reward functions are most effective in which contexts. To our knowledge, no comprehensive study comparing reward explanation techniques in different types of domains has previously been performed. In addition, no comprehensive way of assessing human reward understanding has been proposed. In this paper, we first outline two overarching categories of reward explanations and a subset of explanation modalities within each category. We then suggest a suite of assessment techniques and metrics for human reward understanding, define a set of axes characterizing domain complexity, and discuss a planned human subject experiment designed to better understand which modalities of reward explanations are most effective in domains of varying complexity. We scope this work to consider human understanding of linear reward functions in particular.

\section{Related Work}
\label{related}
Previous works have studied the efficacy of a variety of explanation techniques through human subject experiments in different settings. For example, \citet{chakraborti2019plan} assessed whether human participants could identify optimal versus satisficing plans after receiving explanations intended to reconcile their models with a robot's. The authors also asked post-hoc Likert scale questions about whether the explanations were helpful and easy to understand and whether participants were satisfied with the explanations. In another experiment, \citet{tabrez2019explanation} asked Likert scale-based questions about the helpfulness, sociability, and intelligence of a robot that provided explanations about its reward function to human participants. \citet{wang2016impact} further assessed human-agent team performance, the percentage of correct human decisions, and a human's understanding of an agent's decisions given explanations related to the different components of the agent's POMDP-based representation of a task. Finally, \citet{lage2019evaluation} measured a person's understanding of explanations of various sizes and that provided different types of information using a combination of assessment techniques including simulation of the system's actions, verification of a system's response, and counterfactual reasoning. 

Other works have compared multiple explanation techniques in the context of a single domain. For example, \citet{anderson2019explaining} compared a person's understanding of an agent's reward function given both saliency maps and decomposed reward bars provided at each decision point in a simple Real-Time Strategy game. They measured the person's reward understanding through the accuracy of the person's predictions of the agent's actions coupled with an open-ended questionnaire asking participants to describe the agent's approach or method for making decisions. In another study, \citet{huang2019enabling} assessed a person's reward understanding by asking them to identify an agent's optimal trajectory after being provided with demonstrations of the agent's behavior which were generated either assuming that the human will perform exact inference or approximate inference of the agent's objectives. In their experiment, the human is assumed to know the correct set of features that the agent is using to make decisions. So far, we have not identified any work that has compared a broad set of reward explanation techniques provided through multiple explanation modalities in multiple domains in an experimental setting with human subjects. Further, while some preliminary assessment techniques for human reward understanding have been applied in a subset of these studies, no comprehensive way of assessing human reward understanding exists.

\section{Reward Explanation Techniques}
\label{xaitechniques}
We group reward explanation techniques into two categories: feature space techniques and policy space techniques. Feature space techniques explain the reward function in terms of the individual features that comprise the reward function and their relative weights. Policy space techniques explain the reward function through demonstrations of actions in the environment along with how the demonstrated state-action pairs relate to the policy (best/worst actions, important states, etc.). Note that the goal in both cases is to communicate information about the features and their weights and to support understanding of what this means in terms of action in the environment, but the modality of communication differs between the two categories. Feature space techniques may be most applicable when the reward function can be easily represented by a limited number of interpretable features, while policy space techniques may be ideal when reward functions are uninterpretable or otherwise difficult to reason about in terms of translation into actions in the environment. Here we introduce sub-categories of feature and policy space techniques as well as examples from the literature that fall into each category. These categories represent a broad range of common reward explanation modalities. We intend to select one technique from each category for our future human subject experiment.

\subsection{Feature Space Techniques}

\subsubsection{Direct Reward Function}
One straightforward approach to communicating reward information to humans is to show them the reward function, including all features and their weights, directly. While this might be the most direct and complete way of communicating reward information, there are a number of potential shortcomings of this approach. First, if the domain involves a large number of features, it might be difficult for a person to reason over all of these features simultaneously. Second, in cases in which features are uninterpretable to humans (as with deep reinforcement learning), explaining reward information directly may be infeasible. Finally, even if humans are able to understand and reason over all features and their weights, they may not be able to convert this information into an optimal plan or otherwise use the information for their tasks.

\subsubsection{Feature Subset}

Reward information can also be communicated to humans in feature space through subsets of features and their relative weights. This might be a better approach if there are too many features for the human to reason over simultaneously or in order to help the human to focus only on the most important aspects of robot decision-making, for example. Displaying subsets of features has previously been applied in the context of classification tasks, including producing prototypes of different classes \cite{kim2014bayesian} and identifying the optimal feature subset given a budget of information to display \cite{ribeiro2016should}. While these explanation techniques show users subsets of features, they are not directly applied to reward functions, which we study in this work. \citet{tabrez2019explanation} introduce an technique that infers a human's reward function based on their actions and explains expected missing information. While this technique inherently provides humans with a subset of reward features, in this experiment we are interested in cases in which the reward function is explained from scratch. In our assessment, we will apply a similar approach to that introduced by \citet{ribeiro2016should}. 

\subsubsection{Reward Abstractions}
Finally, reward functions can be explained to humans in feature space using abstractions of features and their relative weights. For example, multiple features may be combined to create one feature or high-level concepts can be combined to create an alternate representation of the reward function. This approach may be especially beneficial if there are too many features for a human to reason over simultaneously or if the features are uninterpretable to humans in some way. Previous works have leveraged user-defined interpretable concepts to learn human-understandable representations of the reward function \cite{lage2020human, sreedharan2020bridging}. In our assessment of reward explanation techniques, we intend to pre-define a set of concepts such as those introduced by \citet{lage2020human} and \citet{sreedharan2020bridging} and learn the appropriate weights via regression.

\subsection{Policy Space Techniques}

\subsubsection{Trajectory Demonstrations}
Another way of revealing information about the robot's reward function to human teammates is through trajectory demonstrations. Trajectory demonstrations could include one or multiple sequences of states and actions generated based on the robot's reward function or policy. For example, the robot could demonstrate the optimal trajectory based on its reward function, much as humans are often assumed to do when teaching a robot through learning from demonstration \cite{argall2009survey}. The robot could also provide the most legible trajectory, where a legible trajectory is defined by \citet{dragan2013legibility} as a trajectory that enables an observer to confidently infer the correct robot goal. Finally, the robot could demonstrate the least optimal trajectory based on its reward function in order to illustrate examples of unfavorable state-action pairs. Note that in many cases, there are many optimal or otherwise equivalent trajectories that could be provided to users, and strategies exist to down-select from these multiple possibilities, for example through the selection of maximally informative trajectories \cite{huang2019enabling, lee2021machine, cakmak2012algorithmic}. In our assessment, we will provide users with the most and least optimal trajectories as a simple baseline.

\subsubsection{Policy Summarization}
Policy summarization techniques demonstrate agent behavior in a subset of informative states given different conditions and scenarios \cite{amir2018agent, lage2019exploring}. Such techniques may be beneficial when the state and action space is large and when it might be difficult for a human user to extrapolate important aspects of the agent's behavior based solely on one or a few optimal or legible trajectories. Given a budget of trajectory segments to provide in a summary, the state-action pairs to include in each segment can be determined based on the importance and diversity of states \cite{amir2018highlights}. Multiple definitions of important states have been proposed, including both Q-function-based \cite{amir2018highlights} and entropy-based \cite{huang2018establishing} definitions.  \citet{amir2018highlights} define important states as states from which taking a wrong action can lead to a significant decrease in future rewards, as determined by
the agent’s Q-values. In particular, they consider the most important states to be states in which the difference between the Q-values associated with the best and worst actions is maximized. We will leverage this definition and the approach introduced by \citet{amir2018highlights} in providing policy summaries.

\subsubsection{Factored Policies via Reward Decomposition}
Finally, reward functions can be explained in policy space through factored policies derived from factored rewards \cite{anderson2019explaining, juozapaitis2019explainable}. In these approaches, the reward function is broken into individual components (in a linear reward function, these might correspond to each feature, for example). A decomposed Q-function with components corresponding to each individual reward component is then learned. With this decomposed Q-function, the contribution of each state-action pair to the individual reward components can be displayed. In our assessment, we will display factored Q-function information for different state-action pairs. We will select states for which we will display factored Q-function information by leveraging a  similar approach to the state importance-based selection strategy described by \citet{amir2018highlights}.

\section{Reward Understanding Assessments}

In order to assess a person's reward understanding, we suggest the use of a suite of four different assessment techniques: free response, feature sub-selection, preference elicitation, and best demonstration elicitation. Since a variety of assessment techniques have been applied in the past and there is not currently a standard way of assessing a human's reward understanding, using a suite of techniques will allow us to consider multiple types of human input and compare the results from each. The following are descriptions of the assessment techniques along with the associated metric for each. We also introduce three composite metrics that are based on the four individual metrics proposed here.

\subsection{Free Response (FR)}
For the free response assessment, subjects will be asked how they think the robot makes decisions in each domain. They will be able to provide free-form answers about the factors they think the robot uses in decision-making and how important each factor is relative to the others. Their responses will be coded in a similar way as in previous literate \cite{anderson2019explaining, kim2017collaborative, hoffman2018metrics}, and
the coded response will be used to produce the set of features that the human believes are important, $F_H^{fr}$, as well as the set of pairwise comparisons of their relative weights, $W_H^{fr}$ (e.g. $w_A > w_B$, where $w_i$ is the weight of feature $i$). Given the robot's ground truth set of features, $F_R^{fr}$, and pairwise comparisons of its relative weights, $W_R^{fr}$, the metric for the free response assessment, based on the similarity metrics used by \citet{shah2020interactive}, can be defined as the intersection over union of the set of human features and pairwise rankings and the ground truth set of robot features and pairwise rankings: \[FR = \dfrac{(F_H^{fr} \cup W_H^{fr}) \cap (F_R^{fr} \cup W_R^{fr})}{(F_H^{fr} \cup W_H^{fr}) \cup (F_R^{fr} \cup W_R^{fr})}\]

\subsection{Feature Sub-selection (FS)}
For the feature  sub-selection  assessment,  subjects  will be provided  with a list of possible features (only a subset of which are actually used by the robot), and they will be asked to select the ones they believe are relevant to the scenario and assign relative weights to each. $F_H^{fs}$, $W_H^{fs}$, $F_R^{fs}$, and $W_R^{fs}$ are defined as in the previous section and yield the following metric for feature sub-selection (also based on the previously-discussed similarity metric): \[FS = \dfrac{(F_H^{fs} \cup W_H^{fs}) \cap (F_R^{fs} \cup W_R^{fs})}{(F_H^{fs} \cup W_H^{fs}) \cup (F_R^{fs} \cup W_R^{fs})}\]

\subsection{Preference Elicitation (PE)}
Preference elicitation involves presenting subjects with multiple trajectories and asking them to select the best, similar to an active learning approach (Settles 2012). We generate queries using the maximum information gain strategy proposed by \citet{biyik2019asking}. We define the set of human responses to these queries as $q_H$ and the set of ground truth correct responses from the robot as $q_R$. Given these, the metric for preference elicitation is defined as the percent of correct human responses (i.e. recall): \[PE = \dfrac{|q_H \cap q_R|}{|q_R|}\]

\subsection{Best Demonstration (BD)}
For the best demonstration assessment, subjects will be asked to provide demonstrations that they believe are optimal given what they know about the agent’s reward function. This assessment is similar to the ``simulation'' assessment used by \citet{lage2019evaluation}. The metric we consider for the best demonstration is the complement of the normalized regret: \[BD = 1 - \dfrac{R(\xi^{*}) - R(\xi^{H})}{R(\xi^{*})} \] $R(\xi^{*})$ is the reward for the optimal trajectory $\xi^{*}$ and $R(\xi^{H})$ is the reward for the human's demonstration $\xi^{H}$. We normalize regret in this case, because all of the other assessment metrics are normalized, and we take the complement, since larger values indicate better understanding for the other assessments. These two steps allowed us to combine the four assessment metrics into composite metrics more readily.

\subsection{Composite Metrics}
Finally, we propose three composite metrics based on combinations of four individual metrics. Just as we divided reward explanation techniques into feature space techniques and policy space techniques, we can similarly divide our assessment metrics into feature space metrics, which ask directly about features and their weights ($FR$ and $FS$), and policy space metrics, which ask about the behaviors that result from reward functions ($PE$ and $BD$). All four individual metrics are normalized, and so we weight them equally in combining them to form the composite metrics. Accordingly, we propose the composite feature space metric: $F = FR + FS$
We also propose the composite policy space metric: $P = PE + BD$
Finally, we propose the overall composite metric: $C = F + P$

\section{Axes of Domain Complexity}
In characterizing domains, we consider four different axes of complexity: reward function complexity, feature complexity, environment complexity, and situational complexity. When considering reward functions that are linear in features, we can vary reward function complexity by considering reward functions with more or fewer features. Feature complexity is related to how complex each individual feature within the linear reward function is. While there might be many ways to characterize feature complexity, we consider the interpretability of individual features as a measure of their complexity. Environment complexity includes factors such as the size of the state and action spaces, whether the state and action spaces are discrete or continuous, or whether a domain is Markovian or non-Markovian. Finally, situational complexity indicates whether a person will need to perform other tasks at the same time as receiving the explanation and the number and difficulty of those tasks.

\section{Proposed Experiment}
We propose an experiment to test each of the different explanation techniques in domains of varying complexity as defined by the four axes outlined in the previous section. We have selected four domains in order to cover a broad range of complexities. Our domains include a simple grid world scenario, 
OpenAI Gym's Lunar Lander 
\cite{brockman2016openai}, the threats and waypoints domain proposed by \citet{shah2018bayesian},  
and the threats and waypoints domain combined with a secondary task in which the human needs to monitor a robot traversing in rocky terrain. 

\subsection{Hypotheses}
Our hypotheses for the proposed experiment include those listed below. We intend to assess our hypotheses using the proposed metrics for reward understanding.

\begin{hyp}
Feature space techniques will lead to better reward understanding than policy space techniques in domains of low versus high reward, feature, and environment complexity. 
\end{hyp}

\begin{hyp}
Policy space techniques will lead to better reward understanding than feature space techniques in domains of high versus low reward, feature, and environment complexity. 
\end{hyp}

\begin{hyp}
The best modality of information (feature versus policy space) will not change between scenarios with low versus high situational complexity in domains of the same reward, feature, and environment complexities.
\end{hyp}

\begin{hyp}
Reward understanding will be worse in scenarios with high versus low situational complexity for both feature space techniques and policy space techniques. 
\end{hyp}

\section{Conclusion}
\label{conclusion}
In this paper, we define categories of existing reward explanation techniques representing a broad set of explanation modalities, and we identify a specific approach we plan to implement in the context of a human subject experiment from each category. We also suggest a suite of assessment techniques and metrics for human reward understanding. These techniques and metrics integrate multiple modalities of human information understanding, including both feature-based information and behavior-/policy-based information. Finally, we define four axes of domain complexity and outline a future experiment to better understand which reward explanation techniques are most effective in which contexts. We hope that the proposed characterization of reward explanation techniques along with the assessment techniques and metrics will contribute to a more systematic understanding of which reward explanation techniques are most beneficial in different contexts through future human subject experiments.

\newpage

\bibliography{bibliography.bib}

\begin{thebibliography}{28}
\providecommand{\natexlab}[1]{#1}
\providecommand{\url}[1]{\texttt{#1}}
\providecommand{\urlprefix}{URL }
\expandafter\ifx\csname urlstyle\endcsname\relax
  \providecommand{\doi}[1]{doi:\discretionary{}{}{}#1}\else
  \providecommand{\doi}{doi:\discretionary{}{}{}\begingroup
  \urlstyle{rm}\Url}\fi

\bibitem[{Amir and Amir(2018)}]{amir2018highlights}
Amir, D.; and Amir, O. 2018.
\newblock Highlights: Summarizing agent behavior to people.
\newblock In \emph{Proceedings of the 17th International Conference on
  Autonomous Agents and MultiAgent Systems}, 1168--1176.

\bibitem[{Amir, Doshi-Velez, and Sarne(2018)}]{amir2018agent}
Amir, O.; Doshi-Velez, F.; and Sarne, D. 2018.
\newblock Agent strategy summarization.
\newblock In \emph{Proceedings of the 17th International Conference on
  Autonomous Agents and MultiAgent Systems}, 1203--1207.

\bibitem[{Anderson et~al.(2019)Anderson, Dodge, Sadarangani, Juozapaitis,
  Newman, Irvine, Chattopadhyay, Fern, and Burnett}]{anderson2019explaining}
Anderson, A.; Dodge, J.; Sadarangani, A.; Juozapaitis, Z.; Newman, E.; Irvine,
  J.; Chattopadhyay, S.; Fern, A.; and Burnett, M. 2019.
\newblock Explaining reinforcement learning to mere mortals: An empirical
  study.
\newblock \emph{arXiv preprint arXiv:1903.09708} .

\bibitem[{Argall et~al.(2009)Argall, Chernova, Veloso, and
  Browning}]{argall2009survey}
Argall, B.~D.; Chernova, S.; Veloso, M.; and Browning, B. 2009.
\newblock A survey of robot learning from demonstration.
\newblock \emph{Robotics and autonomous systems} 57(5): 469--483.

\bibitem[{B{\i}y{\i}k et~al.(2019)B{\i}y{\i}k, Palan, Landolfi, Losey, and
  Sadigh}]{biyik2019asking}
B{\i}y{\i}k, E.; Palan, M.; Landolfi, N.~C.; Losey, D.~P.; and Sadigh, D. 2019.
\newblock Asking easy questions: A user-friendly approach to active reward
  learning.
\newblock \emph{arXiv preprint arXiv:1910.04365} .

\bibitem[{Brockman et~al.(2016)Brockman, Cheung, Pettersson, Schneider,
  Schulman, Tang, and Zaremba}]{brockman2016openai}
Brockman, G.; Cheung, V.; Pettersson, L.; Schneider, J.; Schulman, J.; Tang,
  J.; and Zaremba, W. 2016.
\newblock Openai gym.
\newblock \emph{arXiv preprint arXiv:1606.01540} .

\bibitem[{Cakmak and Lopes(2012)}]{cakmak2012algorithmic}
Cakmak, M.; and Lopes, M. 2012.
\newblock Algorithmic and human teaching of sequential decision tasks.
\newblock In \emph{Twenty-Sixth AAAI Conference on Artificial Intelligence}.

\bibitem[{Chakraborti et~al.(2019)Chakraborti, Sreedharan, Grover, and
  Kambhampati}]{chakraborti2019plan}
Chakraborti, T.; Sreedharan, S.; Grover, S.; and Kambhampati, S. 2019.
\newblock Plan Explanations as Model Reconciliation--An Empirical Study.
\newblock In \emph{2019 14th ACM/IEEE International Conference on Human-Robot
  Interaction (HRI)}, 258--266. IEEE.

\bibitem[{Dragan, Lee, and Srinivasa(2013)}]{dragan2013legibility}
Dragan, A.~D.; Lee, K.~C.; and Srinivasa, S.~S. 2013.
\newblock Legibility and predictability of robot motion.
\newblock In \emph{2013 8th ACM/IEEE International Conference on Human-Robot
  Interaction (HRI)}, 301--308. IEEE.

\bibitem[{Fisac et~al.(2020)Fisac, Gates, Hamrick, Liu, Hadfield-Menell,
  Palaniappan, Malik, Sastry, Griffiths, and Dragan}]{fisac2020pragmatic}
Fisac, J.~F.; Gates, M.~A.; Hamrick, J.~B.; Liu, C.; Hadfield-Menell, D.;
  Palaniappan, M.; Malik, D.; Sastry, S.~S.; Griffiths, T.~L.; and Dragan,
  A.~D. 2020.
\newblock Pragmatic-pedagogic value alignment.
\newblock In \emph{Robotics Research}, 49--57. Springer.

\bibitem[{Hadfield-Menell et~al.(2016)Hadfield-Menell, Dragan, Abbeel, and
  Russell}]{hadfield2016cooperative}
Hadfield-Menell, D.; Dragan, A.; Abbeel, P.; and Russell, S. 2016.
\newblock Cooperative inverse reinforcement learning.
\newblock \emph{arXiv preprint arXiv:1606.03137} .

\bibitem[{Hoffman et~al.(2018)Hoffman, Mueller, Klein, and
  Litman}]{hoffman2018metrics}
Hoffman, R.~R.; Mueller, S.~T.; Klein, G.; and Litman, J. 2018.
\newblock Metrics for explainable AI: Challenges and prospects.
\newblock \emph{arXiv preprint arXiv:1812.04608} .

\bibitem[{Huang et~al.(2018)Huang, Bhatia, Abbeel, and
  Dragan}]{huang2018establishing}
Huang, S.~H.; Bhatia, K.; Abbeel, P.; and Dragan, A.~D. 2018.
\newblock Establishing appropriate trust via critical states.
\newblock In \emph{2018 IEEE/RSJ International Conference on Intelligent Robots
  and Systems (IROS)}, 3929--3936. IEEE.

\bibitem[{Huang et~al.(2019)Huang, Held, Abbeel, and
  Dragan}]{huang2019enabling}
Huang, S.~H.; Held, D.; Abbeel, P.; and Dragan, A.~D. 2019.
\newblock Enabling robots to communicate their objectives.
\newblock \emph{Autonomous Robots} 43(2): 309--326.

\bibitem[{Juozapaitis et~al.(2019)Juozapaitis, Koul, Fern, Erwig, and
  Doshi-Velez}]{juozapaitis2019explainable}
Juozapaitis, Z.; Koul, A.; Fern, A.; Erwig, M.; and Doshi-Velez, F. 2019.
\newblock Explainable reinforcement learning via reward decomposition.
\newblock In \emph{IJCAI/ECAI Workshop on Explainable Artificial Intelligence}.

\bibitem[{Kim, Rudin, and Shah(2014)}]{kim2014bayesian}
Kim, B.; Rudin, C.; and Shah, J.~A. 2014.
\newblock The bayesian case model: A generative approach for case-based
  reasoning and prototype classification.
\newblock In \emph{Advances in neural information processing systems},
  1952--1960.

\bibitem[{Kim, Banks, and Shah(2017)}]{kim2017collaborative}
Kim, J.; Banks, C.; and Shah, J. 2017.
\newblock Collaborative planning with encoding of users' high-level strategies.
\newblock In \emph{Proceedings of the AAAI Conference on Artificial
  Intelligence}, volume~31.

\bibitem[{Lage et~al.(2019{\natexlab{a}})Lage, Chen, He, Narayanan, Kim,
  Gershman, and Doshi-Velez}]{lage2019evaluation}
Lage, I.; Chen, E.; He, J.; Narayanan, M.; Kim, B.; Gershman, S.; and
  Doshi-Velez, F. 2019{\natexlab{a}}.
\newblock An evaluation of the human-interpretability of explanation.
\newblock \emph{arXiv preprint arXiv:1902.00006} .

\bibitem[{Lage and Doshi-Velez(2020)}]{lage2020human}
Lage, I.; and Doshi-Velez, F. 2020.
\newblock Human-in-the-Loop Learning of Interpretable and Intuitive
  Representations.
\newblock In \emph{Proceedings of the ICML Workshop on Human Interpretability
  in Machine Learning, Vienna, Austria}, volume~17.

\bibitem[{Lage et~al.(2019{\natexlab{b}})Lage, Lifschitz, Doshi-Velez, and
  Amir}]{lage2019exploring}
Lage, I.; Lifschitz, D.; Doshi-Velez, F.; and Amir, O. 2019{\natexlab{b}}.
\newblock Exploring computational user models for agent policy summarization.
\newblock \emph{arXiv preprint arXiv:1905.13271} .

\bibitem[{Lee, Admoni, and Simmons(2021)}]{lee2021machine}
Lee, M.~S.; Admoni, H.; and Simmons, R. 2021.
\newblock Machine Teaching for Human Inverse Reinforcement Learning.
\newblock \emph{Frontiers in Robotics and AI} 8: 188.

\bibitem[{Ribeiro, Singh, and Guestrin(2016)}]{ribeiro2016should}
Ribeiro, M.~T.; Singh, S.; and Guestrin, C. 2016.
\newblock " Why should i trust you?" Explaining the predictions of any
  classifier.
\newblock In \emph{Proceedings of the 22nd ACM SIGKDD international conference
  on knowledge discovery and data mining}, 1135--1144.

\bibitem[{Sanneman and Shah(2020)}]{sanneman2020situation}
Sanneman, L.; and Shah, J.~A. 2020.
\newblock A Situation Awareness-Based Framework for Design and Evaluation of
  Explainable AI.
\newblock In \emph{International Workshop on Explainable, Transparent
  Autonomous Agents and Multi-Agent Systems}, 94--110. Springer.

\bibitem[{Shah, Wadhwania, and Shah(2020)}]{shah2020interactive}
Shah, A.; Wadhwania, S.; and Shah, J. 2020.
\newblock Interactive robot training for non-markov tasks.
\newblock \emph{arXiv preprint arXiv:2003.02232} .

\bibitem[{Shah et~al.(2018)Shah, Kamath, Li, and Shah}]{shah2018bayesian}
Shah, A.~J.; Kamath, P.; Li, S.; and Shah, J.~A. 2018.
\newblock Bayesian inference of temporal task specifications from
  demonstrations .

\bibitem[{Sreedharan et~al.(2020)Sreedharan, Soni, Verma, Srivastava, and
  Kambhampati}]{sreedharan2020bridging}
Sreedharan, S.; Soni, U.; Verma, M.; Srivastava, S.; and Kambhampati, S. 2020.
\newblock Bridging the Gap: Providing Post-Hoc Symbolic Explanations for
  Sequential Decision-Making Problems with Black Box Simulators.
\newblock \emph{arXiv preprint arXiv:2002.01080} .

\bibitem[{Tabrez, Agrawal, and Hayes(2019)}]{tabrez2019explanation}
Tabrez, A.; Agrawal, S.; and Hayes, B. 2019.
\newblock Explanation-based reward coaching to improve human performance via
  reinforcement learning.
\newblock In \emph{2019 14th ACM/IEEE International Conference on Human-Robot
  Interaction (HRI)}, 249--257. IEEE.

\bibitem[{Wang, Pynadath, and Hill(2016)}]{wang2016impact}
Wang, N.; Pynadath, D.~V.; and Hill, S.~G. 2016.
\newblock The impact of pomdp-generated explanations on trust and performance
  in human-robot teams.
\newblock In \emph{Proceedings of the 2016 international conference on
  autonomous agents \& multiagent systems}, 997--1005.

\end{thebibliography}

\end{document}